\documentclass[submission,copyright,creativecommons]{eptcs}
\providecommand{\event}{STAI 2023} 

\usepackage{iftex}
\usepackage{graphicx}
\usepackage{hyperref}

\ifpdf
  \usepackage{underscore}         
  \usepackage[T1]{fontenc}        
\else
  \usepackage{breakurl}           
\fi

\title{Analysing Gender Bias in Text-to-Image Models \\ using Object Detection}
\author{Harvey Mannering
\institute{University College London}
\email{harvey.mannering@ucl.ac.uk }}
\def\titlerunning{Analysing Gender Bias in Text-to-Image Models using Object Detection}
\def\authorrunning{H. Mannering}
\begin{document}
\maketitle
\begin{sloppypar}

\begin{abstract}
This work presents a novel strategy to measure bias in text-to-image models.  Using paired prompts that specify gender and vaguely reference an object (e.g. “\texttt{a man/woman holding an item}”) we can examine whether certain objects are associated with a certain gender.  In analysing results from Stable Diffusion, we observed that male prompts generated objects such as ties, knives, trucks, baseball bats, and bicycles more frequently. On the other hand, female prompts were more likely to generate objects such as handbags, umbrellas, bowls, bottles, and cups. We hope that the method outlined here will be a useful tool for examining bias in text-to-image models.
\end{abstract}

\section{Introduction}

Text-to-image models are neural networks that take as input a text prompt and output an image. This has exciting new applications for storyboarding, image editing, and AI-generated art; however, it also poses risks. These models are capable of depicting stereotypes that have been learned from the training data.  For instance, DALL·E, Stable Diffusion, and Midjourney were more prone to producing images of men if the word ``\texttt{powerful}'' was included in the text prompt \cite{fraser2023friendly}.

While some research in this field currently uses gender-neutral prompts (e.g. “\texttt{a photo of a nurse}”) to examine what gender things are associated with, we take the reverse approach.  We use prompts with a specified gender and vague references to objects (e.g. “\texttt{a girl holding an item}”).Object detection can then be used to determine if certain objects are associated with a specific gender. We hope this method will be helpful in analysing biases held by a specific model. For code, prompts, and results, please visit our \href{https://github.com/harveymannering/Text-to-Image-Bias}{Github repo}.

\begin{figure}[!ht]
\centering
\begin{tabular}{cc|cc}   
\multicolumn{2}{c|}{(a) Stable Diffusion} & \multicolumn{2}{c}{(b) DALL·E mini} \\
\multicolumn{2}{c|}{ } & \multicolumn{2}{c}{ } \\
  \includegraphics[width=0.22\linewidth]{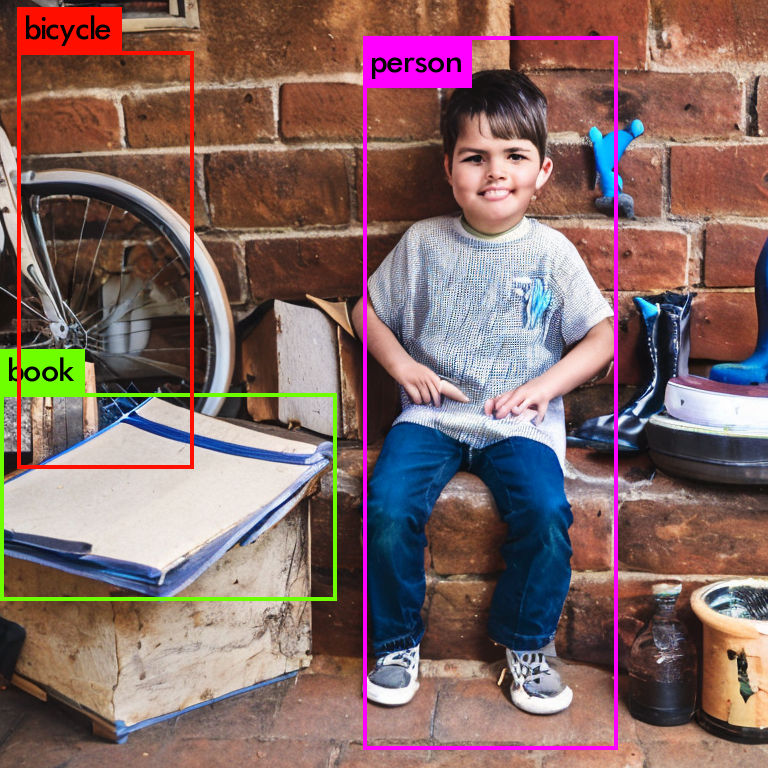} &
  \includegraphics[width=0.22\linewidth]{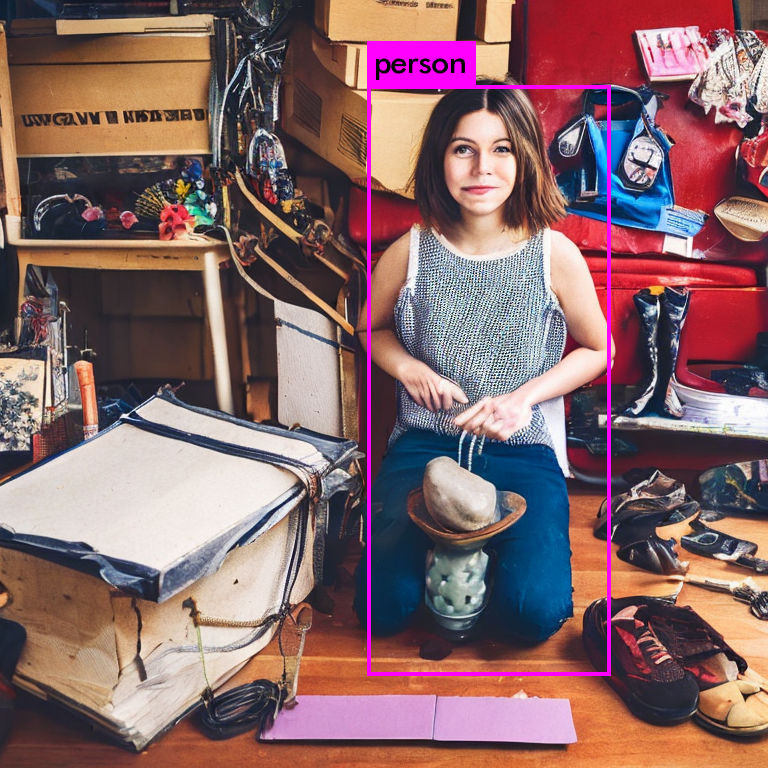} &
  \includegraphics[width=0.22\linewidth]{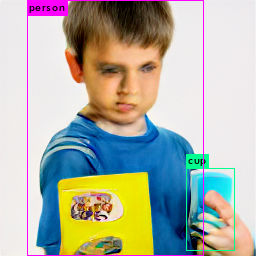} &
  \includegraphics[width=0.22\linewidth]{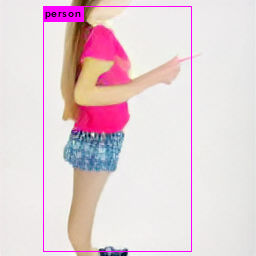} \\
   \footnotesize\texttt{A photo of a boy} & \footnotesize\texttt{A photo of a girl} & \footnotesize\texttt{A photo of a boy} & \footnotesize\texttt{A photo of a girl} \\
   \footnotesize\texttt{with things they own} & \footnotesize\texttt{with things they own} &   \footnotesize\texttt{with an object} & \footnotesize\texttt{with an object}\\
\end{tabular}
\caption{Example outputs from (a) \textbf{Stable Diffusion}  and (b) \textbf{DALL·E mini} models with their corresponding prompts.  Objects detected using YOLOv3 are also shown.} \label{SDExamples}
\end{figure}

\section{Related Work}

Text-to-image models have recently gained a lot of attention due to their realism and versatility.  DALL·E 2 \cite{ramesh2022hierarchical} (a popular text-to-image model) works in two stages. Firstly, a CLIP embedding is generated from a text caption. CLIP is a deep learning model that connects images and textual descriptions by learning a shared embedding space \cite{radford2021learning}.  In the second stage, a diffusion model generates an image conditioned on the CLIP embedding. Stable diffusion \cite{StableDiffusion} similarly encodes text using CLIP; however, the diffusion process in the second stage is performed in the latent space of a pretrained autoencoder, allowing for faster training and inference times. Text-to-image models are now widely used, with DALL·E alone having over a million users \cite{DALLEWaitlist}.

Naturally, researchers are now probing these models to see what biases they contain.  This has been done in DALL-Eval \cite{cho2022dall}, which used neutral prompts like “\texttt{a photo of nurse}” and “\texttt{a person with a beer}” to generate images.  These images were then analysed using automated gender detection, automated skin detection, and human evaluation.  With this pipeline, they determined whether text-to-image models were perpetuating stereotypes. Bias was demonstrated by Bianchi \textit{et al.} \cite{bianchi2022easily} using the prompt template “\texttt{a photo of a face of \{x\}}”.  Only images of men and only images of women were generated when \texttt{\{x\}} was set to ``\texttt{a software developer}'' and ``\texttt{a flight attendant}'', respectively.  This may be because when a prompt is underspecified (i.e. when few details about a person are given) a text-to-image model is forced to ``fill in the blanks'' with stereotypes learned from the training data \cite{fraser2023friendly}. Stable Bias \cite{luccioni2023stable} examines gender bias by generating a large number of images and then analysing them with captioning and visual question answering models. Two metrics, GEP \cite{zhang2023auditing} and MCAS \cite{mandal2023multimodal}, have recently been proposed to measure gender bias in text-to-image models.  Both use CLIP embeddings to examine what associations men and women have. In our analysis, we instead utilize an object detector and prompt in a more open ended way. 

\section{Method}

To determine what associations men and women have in text-to-image models we generate images using male/female paired prompts.  For example, the following two prompts (1) ``\texttt{A man holding an item}'' (2) ``\texttt{A woman holding an item}'' will generate similar, but distinct, images.  Object detection is then run on the resulting images.  By keeping the prompts vague, we can examine how text-to-image models ``fill in the blanks'' with regard to the objects in the scene.  Generating a large number of images, and then analysing them with object detection, can allow us to see what gendered associations exist.

We use 50 template prompts that all contain a gendered word and some vague underspecified reference to an object.  For example, one of our prompts follows the template: ``\texttt{Things owned by a \{gender\}}''
where \texttt{\{gender\}} is set to either “\texttt{man}'', “\texttt{woman}'', “\texttt{boy}'' or “\texttt{girl}''.  These four gender words along with the 50 templates give us a total of 200 prompts that can be used to generate images.

We generate 1000 images for both Stable Diffusion v2-1 \cite{StableDiffusion} and DALL·E mini \cite{DalleMini}.  This involves generating 5 images for each prompt.  Every time a pair of man/woman or boy/girl prompts are used, the same seed is set.  This ensures that the same noise is used in the diffusion process and that the only thing that changes between generations is the gendered word. We selected these light weight models due to our limitations in cost and computational resources.

For object detection we use the You Only Look Once (YOLO) v3 model \cite{redmon2018yolov3} due to its low compute costs.  YOLOv3 can detect multiple objects within the same image. It also draws a bounding box around each object and assigns a probability to the detection.  Example YOLOv3 predictions can be seen in Figure \ref{SDExamples}.

\begin{figure}[!ht]
\centering
\begin{tabular}{cc}   
  \includegraphics[width=0.87\linewidth]{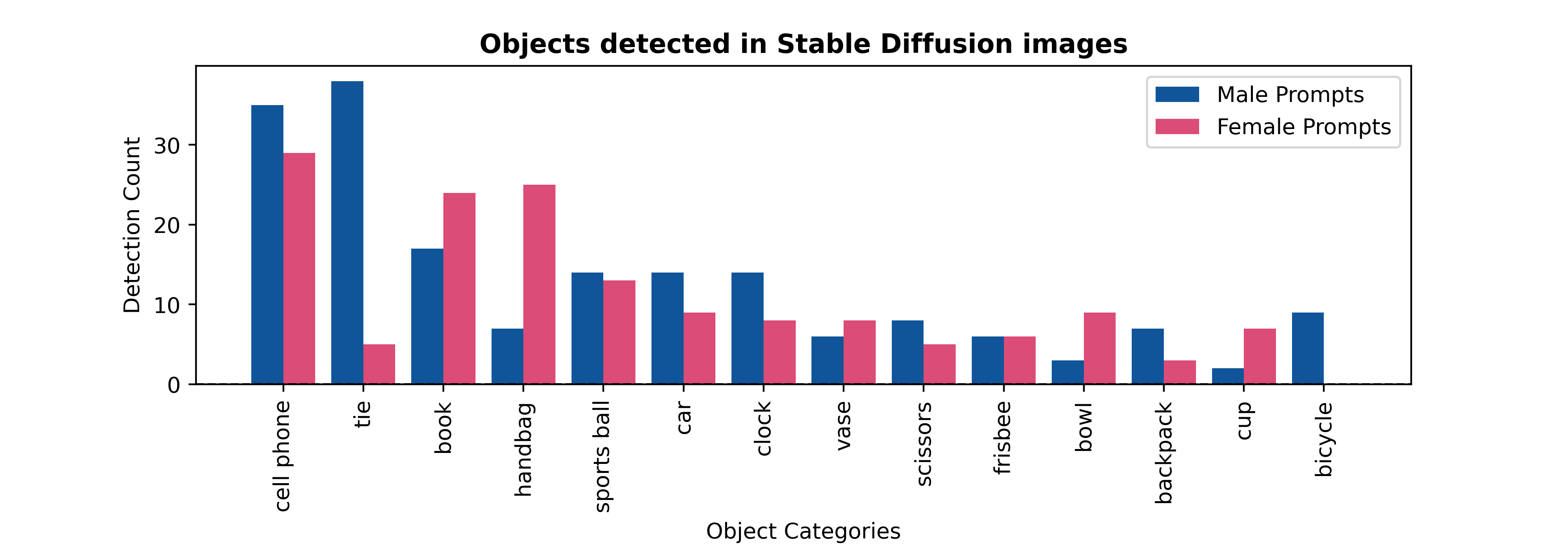} \\
  \includegraphics[width=0.87\linewidth]{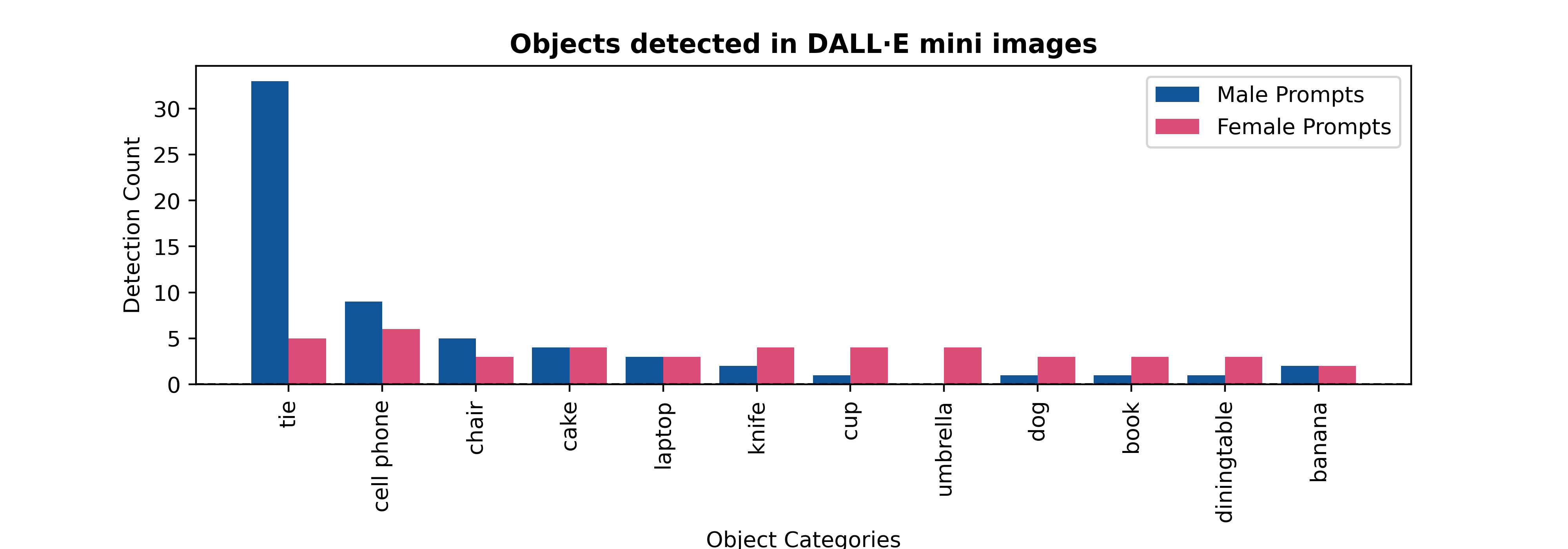} 

\end{tabular}
\caption{Object detection was run on text-to-image generated images.  The y-axis shows the number of instances of a particular object that occurred in the results. Objects are listed on the x-axis. Blue bars correspond to the objects generated from male prompts and the pink bars correspond to objects generated from female prompts. Any object that occurred less than 9 times was removed from the \textbf{Stable diffusion (top)} plot.  Objects with less than 4 occurrences were removed from the \textbf{DALL·E mini (bottom)} plot.  The ``person'' object was removed from both plots.} \label{Results}
\end{figure}

\section{Results}

We generate 1000 images from both Stable Diffusion and DALL·E mini.  We then run object detection on every image.  Examples of the resulting images and detected objects are shown in Figure \ref{SDExamples}. Figure \ref{Results} shows which objects were detected, and in what quantity, for both male and female prompts.  Only the most numerically significant objects are shown, but the full list of objects can be seen in Table~\ref{results-table}.

For Stable Diffusion, male prompts were more likely to generate objects including {ties}, {backpacks}, {knives}, {trucks}, {chairs}, {baseball bats}, and {bicycles}.  Female prompts were more likely to generate objects including {handbags}, {umbrellas}, {bowls}, {bottles}, and {cups}.  Many objects in DALL·E mini were ambiguous, leading to far fewer objects being picked up during object detection. Like with Stable Diffusion, the most significant result is that ties were much more likely to be generated by male prompts.

For each model, we have a male and a female categorical distribution. These two categorical distributions can be compared using the Chi-squared test. The Chi-squared test’s p-value describes how similar the two distributions are, and we can therefore use it as a measure of gender bias (with a higher number meaning less bias). A Chi-squared test performed on the Stable Diffusion gets a p-value of 0.000009. A Chi-squared test for DALL·E mini nets a p-value of 0.04172. This suggests that DALL·E mini contains less gender bias than Stable Diffusion, which may be explained by steps taken by OpenAI to reduce bias in DALL·E \cite{DALLEMitigation}. However, this may also be down to fewer objects being detected in DALL·E mini’s results.

\begin{table}[!ht]
\centering
\parbox{.355\linewidth}{
\tiny
\centering
\begin{tabular}{ccccc}
\hline
 & \multicolumn{2}{c}{\textbf{SD}} & \multicolumn{2}{c}{\textbf{DALL·E}}  \\
 & Male & Female & Male & Female \\
\hline
person & 482 & 515 & 424 & 459\\
sports ball & 14 & 13 & 2 & 1\\
handbag & 7 & 25 & 1 & 1\\
book & 17 & 24 & 1 & 3\\
vase & 6 & 8 & 2 & 0\\
boat & 0 & 1 & 0 & 0\\
donut & 0 & 2 & 0 & 0\\
frisbee & 6 & 6 & 0 & 2\\
baseball glove & 4 & 1 & 1 & 1\\
backpack & 7 & 3 & 0 & 0\\
car & 14 & 9 & 0 & 0\\
umbrella & 2 & 5 & 0 & 4\\
clock & 14 & 8 & 2 & 0\\
cell phone & 35 & 29 & 9 & 6\\
orange & 2 & 4 & 1 & 0\\
diningtable & 1 & 3 & 1 & 3\\
pizza & 0 & 2 & 0 & 0\\
bed & 2 & 4 & 2 & 1\\
pottedplant & 0 & 3 & 0 & 1\\
truck & 4 & 1 & 0 & 0\\
toothbrush & 0 & 4 & 1 & 0\\
mouse & 1 & 3 & 0 & 0\\
knife & 6 & 2 & 2 & 4\\
skateboard & 0 & 1 & 2 & 0\\
tvmonitor & 3 & 3 & 0 & 1\\
bowl & 3 & 9 & 0 & 0\\
bench & 2 & 1 & 0 & 0\\
surfboard & 0 & 1 & 1 & 0\\
bottle & 1 & 5 & 2 & 1\\
teddy bear & 2 & 3 & 2 & 1\\
fork & 0 & 1 & 0 & 0\\
\hline
\\
\end{tabular}
} 
\parbox{.35\linewidth}{
\tiny
\centering
\begin{tabular}{ccccc}
\hline
 & \multicolumn{2}{c}{\textbf{SD}} & \multicolumn{2}{c}{\textbf{DALL·E}}  \\
 & Male & Female & Male & Female \\
\hline
cup & 2 & 7 & 1 & 4\\
tie & 38 & 5 & 33 & 5\\
cake & 1 & 2 & 4 & 4\\
toilet & 0 & 1 & 0 & 0\\
laptop & 1 & 1 & 3 & 3\\
cat & 2 & 1 & 2 & 1\\
scissors & 8 & 5 & 1 & 1\\
spoon & 2 & 1 & 0 & 0\\
baseball bat & 7 & 1 & 1 & 0\\
bird & 0 & 1 & 0 & 0\\
chair & 5 & 1 & 5 & 3\\
hot dog & 0 & 2 & 0 & 0\\
wine glass & 1 & 1 & 0 & 0\\
suitcase & 4 & 3 & 2 & 1\\
microwave & 0 & 1 & 0 & 0\\
apple & 2 & 1 & 0 & 0\\
bicycle & 9 & 0 & 0 & 0\\
dog & 2 & 0 & 1 & 3\\
remote & 1 & 0 & 2 & 0\\
motorbike & 2 & 0 & 0 & 0\\
banana & 1 & 0 & 2 & 2\\
train & 0 & 0 & 0 & 1\\
refrigerator & 0 & 0 & 2 & 1\\
elephant & 0 & 0 & 1 & 1\\
carrot & 0 & 0 & 1 & 1\\
bear & 0 & 0 & 0 & 1\\
zebra & 0 & 0 & 2 & 0\\
tennis racket & 0 & 0 & 1 & 0\\
oven & 0 & 0 & 1 & 0\\
stop sign & 0 & 0 & 1 & 0\\ \\
\hline
\\
\end{tabular}

} \caption{\label{results-table} Total objects in the images generated by Stable Diffusion (SD) and DALL·E mini (DALL·E), divided by the gender used for the input prompt.}
\end{table} 

\section{Conclusion \& Future Work}

In this work, we propose a new technique for measuring bias in text-to-image models.  Using prompts describing a male or a female with an unspecified object, and then running object detection on the results, we can analyse what associations text-to-image models hold about males and females.   This technique opens avenues for further investigation into bias in these models. For instance, our findings indicate that Stable Diffusion tends to generate bowls, bottles, and cups more frequently for women than men. This prompts us to question whether Stable Diffusion portrays women in domestic settings more often than men, thus reinforcing certain stereotypes.  A likely cause of bias in text-to-image models is bias being present in the training data. Therefore, better curation of this data may be needed to address the issue.

These experiments are an initial step towards addressing gender bias. Future work should also include non-binary or trans categories. A good first step in this direction would be to expand the gendered words to ``\texttt{man}'', ``\texttt{woman}'', and ``\texttt{person}''.  Future work could also look to measure other biases (e.g. racial, age, social) using the same technique. Finally, the object detectors own biases needs closer examination. Could YOLO be missing certain detections because they are paired with men or women? The choice of categories used to train YOLO could also influence the outcomes and interpretations of the experiments.

\end{sloppypar}

\bibliographystyle{eptcs}
\bibliography{example}
\end{document}